\documentclass{bmvc2k}

\usepackage{subcaption}
\usepackage{amssymb}
\usepackage{paralist}

\DeclareGraphicsExtensions{.pdf,.jpg,.jpeg,.png}
\graphicspath{{figs/}}

\newcommand{\seclab}[1]{\label{sec:#1}}

\newcommand{\figlab}[1]{\label{fig:#1}}
\newcommand{\figref}[1]{Figure~\ref{fig:#1}}

\title{Learning Geo-Temporal Image Features}

\newcommand{\authoritem}[2]{#1$^#2$ }
\newcommand{\emailitem}[2]{ {\tt\small #1@#2} }
\newcommand{\institutionitem}[2]{ $^#2$#1 }

\author{
  \begin{minipage}{\linewidth}
    \centering
    \authoritem{Menghua Zhai}{1}
    \authoritem{Tawfiq Salem}{1}
    \authoritem{Connor Greenwell}{1}\\
    \authoritem{Scott Workman}{1}
    \authoritem{Robert Pless}{2}
    \authoritem{Nathan Jacobs}{1}
    \\[.10cm]
    \institutionitem{University of Kentucky}{1}
    \hfill
    \institutionitem{George Washington University}{2}
    \\[.10cm]
    \emailitem{\{ted,salem,connor,scott,jacobs\}}{cs.uky.edu}
    \hfill
    \emailitem{pless}{gwu.edu}
  \end{minipage}
}

\runninghead{MZ, TS, CG, SW, RP, NJ}{Learning Geo-Temporal Image Features}

\begin{document}

\maketitle

\begin{abstract}

  We propose to implicitly learn to extract geo-temporal image
  features, which are mid-level features related to when and where an
  image was captured, by explicitly optimizing for a set of location
  and time estimation tasks. To train our method, we take advantage of
  a large image dataset, captured by outdoor webcams and cell phones.
  The only form of supervision we provide are the known capture time
  and location of each image. We find that our approach learns
  features that are related to natural appearance changes in outdoor
  scenes. Additionally, we demonstrate the application of these
  geo-temporal features to time and location estimation. 

\end{abstract}

\section{Introduction}

Outdoor images often contain sufficient visual information to
understand geographic information about the scene, such as where the
image was captured. Developing effective algorithms for this task has
received significant attention for many years~\cite{im2gps,weyand2016planet}.
The appearance of an outdoor scene can also change rapidly. These
changes are often due to fleeting, or transient, attributes such as
lighting and weather conditions, that dramatically affect the visual
perception of an environment. For instance, consider a scene that
changes from sunny and pleasant to rainy and brooding in mere minutes.
Several methods have been proposed for automatically understanding and
extracting these subtle characteristics from
imagery~\cite{patterson2012sun,lu2014two,laffont2014transient,baltenberger16transient}.
Estimating these types of transient attributes has importance in a
number of applications, including: environmental
monitoring~\cite{wang2013observing,fedorov2014snow}, as a
pre-processing step for
calibration~\cite{jacobs13cloudcalibration,workman2014rainbow}, and
enabling semantic browsing of large photo
collections~\cite{jacobs07amos,laffont2014transient}. Our work fuses these two
research areas by learning to estimate geo-temporal image features,
which are related to when and where an image was captured.
 
Recently, a significant amount of work has explored how sources of
supervision beyond manual annotation can be used to learn useful
representations of images. In general, collecting manual annotations
for millions, or perhaps billions, of images is prohibitively
expensive. As Doersch summarizes~\cite{doersch2016supervision}, ``The
idea is that, given the right task, the computer can learn on its own
to represent useful semantic properties of the visual world.'' Such
learning tasks are often referred to as {\em pretext tasks}; they
serve as an intermediary target for learning the intended
representation.  For example, Doersch et
al.~\cite{doersch2015unsupervised} show how spatial context can be used
as a supervisory signal in order to learn a visual representation for
object discovery. Similarly, Pathak et al.~\cite{pathak2016context}
use context-based pixel prediction for pre-training a representation
for classification, detection, and segmentation tasks. We extend this
line of work by using time and location context to learn useful
features from a large corpus of imagery.

Our work makes the following visual assumptions about the world.
First, that photographs provide a direct source of context regarding
the conditions under which they were captured. For example, the time
of day that an image is captured is directly related to the brightness
of the image (i.e., light to dark), season can indicate the expected
weather conditions or how people are dressed, and location can provide
evidence about anticipated styles, such as architecture. Second, these
context signals are hard to extract from an image, are potentially
noisy (e.g., snow in early Summer), and can be indicated by multiple sources
(e.g., snow on the ground, people wearing heavy coats). These assumptions
motivate our method which integrates image appearance, time, and
location, the latter of which are typically recorded automatically by
the imaging device.

In our approach, we explicitly model the relationship between the
image, its geographic location, and the time of capture. We propose a
novel convolutional neural network architecture that implicitly learns
how to extract geo-temporal features from the imagery by optimizing for
a set of location and time estimation tasks. Specifically,
we structure our network to jointly learn feature representations for
three related spaces: images, time, and location. To accomplish this,
each representation, or combination of representations, is used to
predict held out information. For example, the image representation
and location representation (or the combination of both) are used to
learn to predict when an image was captured. In total, three representations 
are learned using four classification tasks.  We optimize all
representations and tasks simultaneously, in an end-to-end fashion.

The main contributions of this work are: 1) a novel approach for
learning geo-temporal image features from a large corpus of imagery
without requiring image-level manual annotations; 2) an evaluation of
the learned features on the task of transient attribute estimation,
where our features outperform those from a network pre-trained using
the strongly supervised ImageNet dataset~\cite{ILSVRC15}; 3) an
evaluation of the accuracy of our learned estimators, highlighting the
value of additional context; and 4) a novel location estimation method
that uses the task of time estimation to localize a static webcam. 

\section{Related Work}

Image localization, or estimating where an image was captured, is an
important problem in the vision community. Typically, the problem is
formulated as image retrieval using a reference database of
ground-level images~\cite{im2gps} or overhead
images~\cite{lin2015learning,workman2015geocnn,workman2015localize}
with known location.  Other methods have been proposed which take
advantage of photometric and geometric properties such as sun
position~\cite{lalonde2010sun,workman2014rainbow}, and many other
cues. More recently, Weyand et al.~\cite{weyand2016planet} proposed to directly
predict the geographic location of a single image using a deep
convolutional neural network by classifying the query image into a set
of spatial bins. For our localization task, we adopt this
classification approach and extend it to include temporal context.

Other work has explored how to estimate the time that an image was
captured. Salem et al.~\cite{salem2016dating} demonstrate that human
appearance, including clothing and hairstyle, is a useful cue for
dating images. Matzen and Snavely~\cite{matzen2014scene} predict
timestamps for photos by matching against a time-varying
reconstruction of a scene.  Volokitin et al.~\cite{eth_biwi_01292} use
representations extracted from CNNs to estimate ambient temperature
and time of year for outdoor images. As with localization, we adopt a
classification approach to estimating when an image was captured and
show how these estimates improve when the image location is known.

Attribute-based representations have become popular in outdoor scene
understanding to help describe how the appearance of a scene changes
over time. Laffont et al.~\cite{laffont2014transient} introduced a taxonomy
of 40 transient attributes that describe intra-scene variations along
with methods for identifying the presence of such attributes in an
image. Using this dataset, Baltenberger et
al.~\cite{baltenberger16transient} introduce methods for estimating
the presence of transient attributes using convolutional neural
networks. Jacobs et al.~\cite{jacobs07amos} demonstrate
that principal component analysis, when applied to webcam imagery,
results in a decomposition that is closely related to natural changes
in the scene, including the time of day, local weather conditions, and
human activity. Similarly, a body of work has sought understand local
weather conditions~\cite{islam13webcamweather,lu2014two}. Many studies
have shown that these types of transient attributes can be useful for
image and camera localization
tasks~\cite{jacobs07geolocate,baltenberger16transient}.

Recent work has explored the use of self-supervision, which are sometimes
referred to as pretext tasks, for training
deep neural networks to capture useful visual
representations~\cite{doersch2015unsupervised,pathak2016context}. For
example, Zhang et al.~\cite{zhang2016colorful} show how image
colorization (synthesizing colors for a grayscale image) is a powerful
pretext task for learning visual representations. Pathak et
al.~\cite{pathak2017learning} exploit low-level motion-based grouping
cues for unsupervised feature learning.
These methods typically
exploit some known quantity of the data (e.g., pixel color values) to
avoid expensive manual annotation. As a byproduct, a useful visual
representation is learned. In our work, we consider two novel pretext
tasks, time and location estimation. 

\section{Estimating Geo-Temporal Image Features}
\seclab{when-and-where}

\begin{figure}[t]
  \centering

  \includegraphics[width=.7\linewidth]{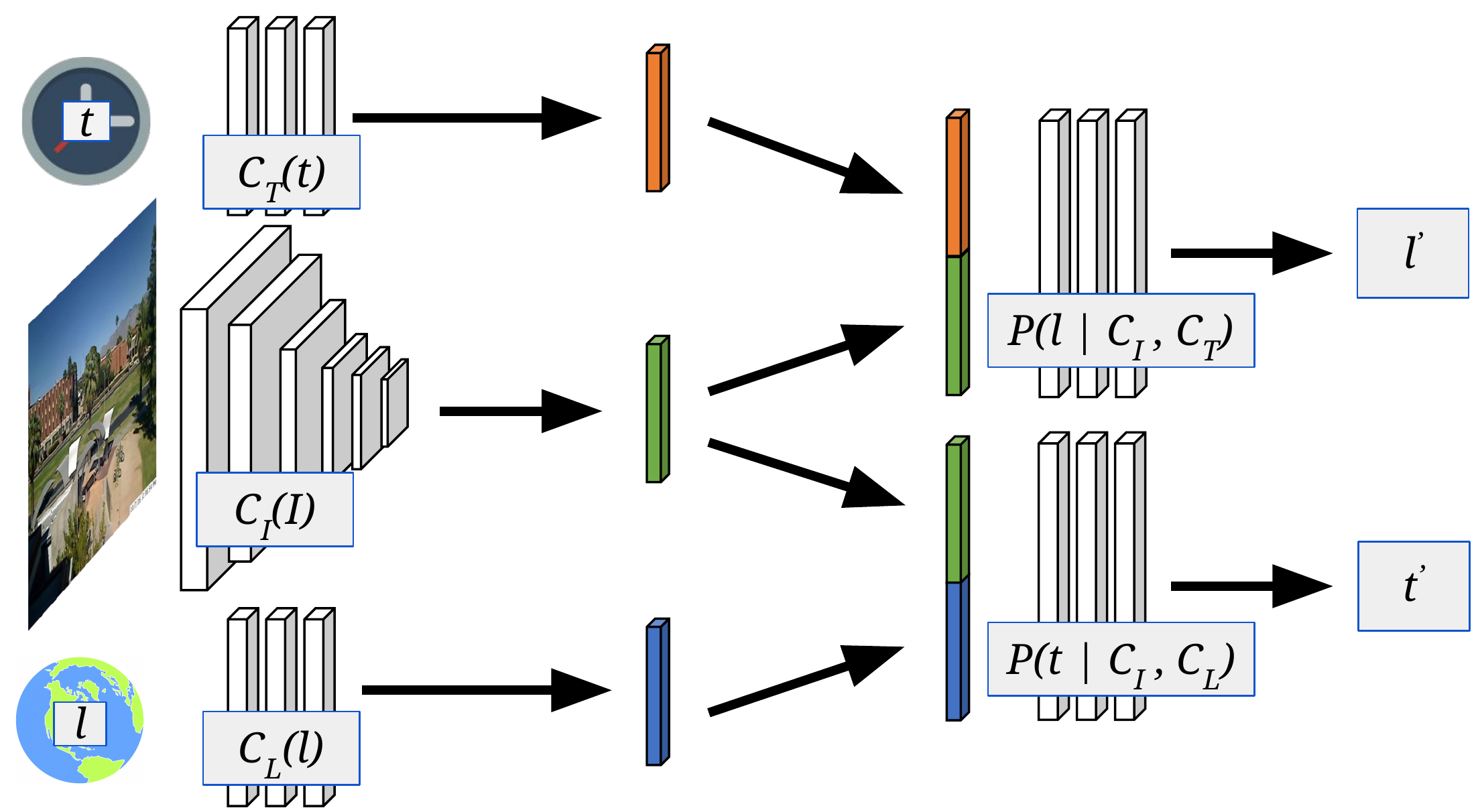} \caption{An
  overview of the proposed network architecture. Our approach learns mid-level
  feature representations for time (orange), location (blue), and
  image appearance (green) by optimizing for a set of conditional time
  and location estimation tasks.}
  
  \figlab{architecture}
\end{figure}

We propose a neural network architecture for learning geo-temporal
features from images by optimizing for a set of
location and time estimation tasks. An overview of the proposed
architecture is shown in \figref{architecture}. Our network takes
three inputs: an image, $I$, the time the image was captured, $t$, and
the location of capture, $l$. Each input is independently processed by
a {\em context network} to extract mid-level features. Then, pairs of
these features are used by {\em estimator networks} to predict
distributions over time or location.

\subsection{Context Networks}

We use three {\em context networks}: a temporal context network,
$C_T(t)$; a location context network, $C_L(l)$; and an image context
network, $C_I(I)$. The output of each context network is a
128-dimensional feature with a sigmoid activation function. 
For the temporal context network, we parameterize the input timestamp
using a one-hot encoding of month and hour of day, for a total of $12
\times 24$ dimensions.  This encoding is flattened and passed to
$C_T(t)$, which consists of three fully-connected layers (with 256,
512, and 128 channels respectively), the first two with ReLU
activations. 
For the location context network, we parameterize the geographic
location, $l$, using standard 3D ECEF coordinates, which we normalize
by the Earth's radius. Other than a different input and independent
network weights, the location context network is identical to the
temporal context network. 
For the image context network, we use the {\em InceptionV2}
architecture~\cite{szegedy2016rethinking}, up to the global
pooling layer, to extract features. We flatten the output
feature map and append the same structure as the other context
networks.

\subsection{Estimator Networks}

The output of the context networks are used as input to four different
estimator networks:
\begin{compactitem}
  
  \item {\em Location Estimator}, $P(l|(C_I(I))$, which predicts
    location using only image features;

  \item {\em Time Estimator}, $P(t|C_I(I))$, which predicts capture time using
    only image features; 

  \item {\em Time-conditioned Location Estimator}, $P(l|(C_I(I),C_T(t))$, which predicts
    location using features from the image and the known capture time;

  \item {\em Location-conditioned Time Estimator}, $P(t|C_I(I),C_L(l))$, which
    predicts capture time using features from the image and the known geographic
    location. 

\end{compactitem}
Aside from different output sizes, the estimator
networks have the same structure as the context networks. We
discretize the output space for location and time and represent the
probability as a categorical distribution (i.e., using a {\em softmax}
activation for each estimator). For location, we use $37 \times 72$
equal-angle ``latitude$\times$ longitude'' bins.  For time, we use $12
\times 24$ ``month $\times$ hour'' bins.

\subsection{Implementation Details}

We randomly initialize the {\em InceptionV2} network using the
standard strategy~\cite{szegedy2016rethinking}.
We initialize all other network weights randomly using Xavier
initializer~\cite{glorot2010understanding} and
simultaneously optimize them during training. For each
estimator network, we have a cross entropy loss.  We minimize the sum
of these using the {\em Adam} optimizer \cite{kingma2014adam}
($\beta_1 = 0.9$ and $\beta_2 = 0.999$). We use a learning rate policy
that starts from 0.001 and decreases by half every $50k$ iterations.
For regularization, we apply weight decay with rate of 0.0001. We
train the proposed network for $2.5M$ iterations with batch size 32.
We apply batch
normalization~\cite{inception15} on every layer except the last (for
both context and estimator networks).  The input images are scaled to
$[-1,1]$ and augmented by a random crop to the size of $224 \times
224$. We use Greenwich Mean Time (GMT) for all timestamps.

\section{Experiments}

We evaluate the context networks and estimator networks on various
datasets, visualize specific features in the image context networks,
and show that the image context features have strong correlations with
transient image attributes.

\subsection{Training and Evaluation Datasets}

We use four main datasets to evaluate our approach.  The {\em AMOS}
dataset refers to a subset of the AMOS
database~\cite{jacobs07amos}, which is a collection of over a billion
images captured from public outdoor webcams around the world.  For our
experiments, we use a subset of images: only from webcams with
high-accuracy geolocation and images captured between 2002 to 2017.
This resulted in images from 12,193 webcams from which we held-out 231
for testing. Each image has a timestamp recorded by the image
collection process.  The {\em YFCC} dataset refers to a subset of
the Yahoo 100 million dataset~\cite{yfcc100m}, only including
geotagged images from smart phones.  We restricted the dataset to
smart phone images since we found that non-phone images often had
inaccurate timestamps.  We filter out indoor images using the {\em
Places} network~\cite{zhou2017places}. This results in a training set
of 892,662 images and a test set of 170,994 images.  The
{\em Hybrid} dataset refers to a combination of the {\em AMOS} and
{\em YFCC} training sets (sampling equally for each mini-batch).  The
{\em TA} dataset refers to the Transient Attributes
Dataset~\cite{laffont2014transient}, which contains 8,571 images, each
manually annotated with $40$ transient attributes, such as sunny and
cloudy.

\subsection{Understanding the Image Context Representation}

\begin{figure*}[t] 
  \centering
  
  \setlength\tabcolsep{2pt}
  \renewcommand{\arraystretch}{0}
  \begin{tabular}{lccc}
    \vspace {0.1cm}
    \raisebox{.3\height}{\rotatebox{90}{\em YFCC}} &
    \includegraphics[width=.31\linewidth]{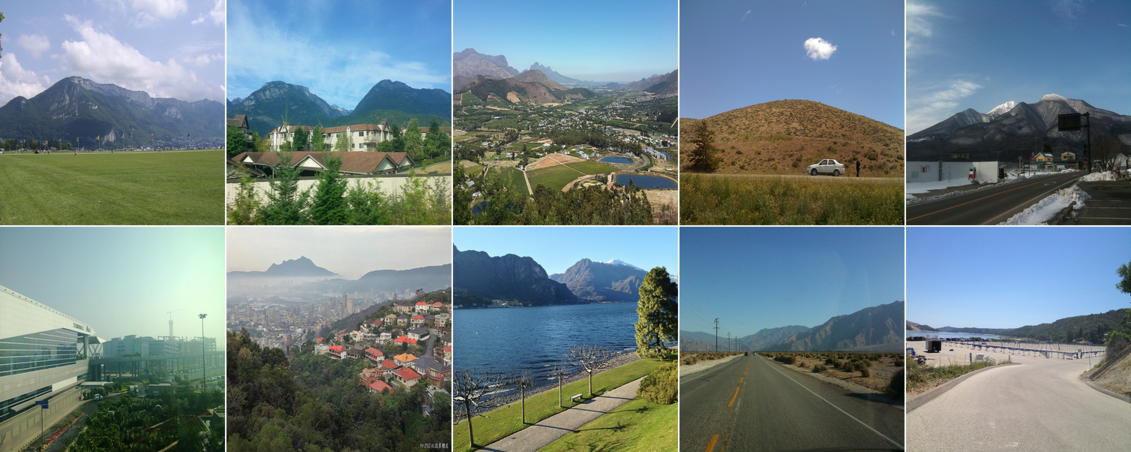} &
    \includegraphics[width=.31\linewidth]{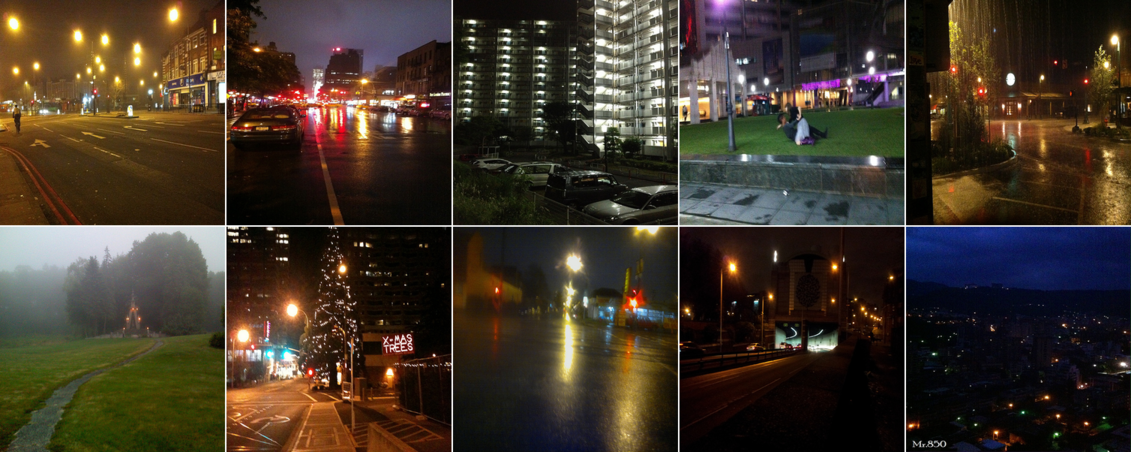}&
    \includegraphics[width=.31\linewidth]{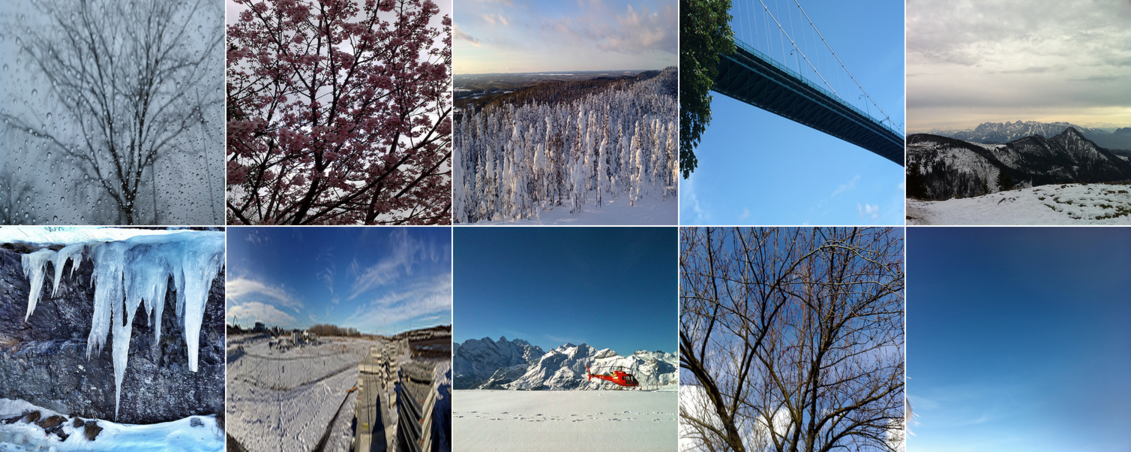}\\
    \raisebox{-.05\height}{\rotatebox{90}{{\em AMOS:4308}}} &
    \includegraphics[width=.31\linewidth]{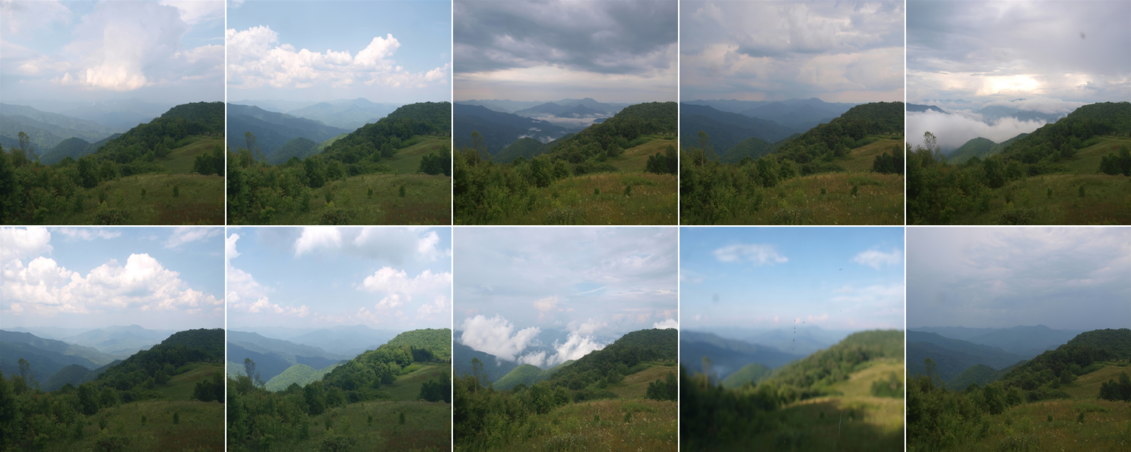} &
    \includegraphics[width=.31\linewidth]{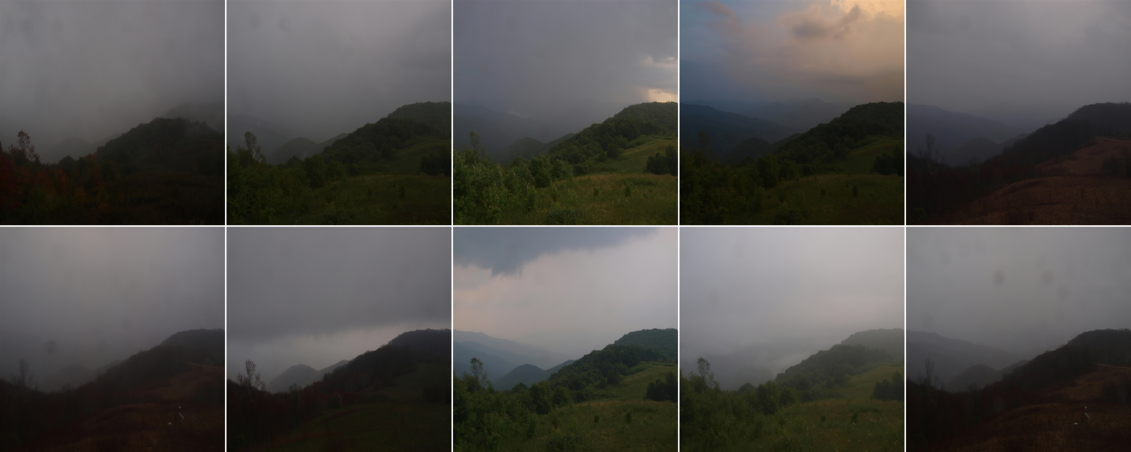} &
    \includegraphics[width=.31\linewidth]{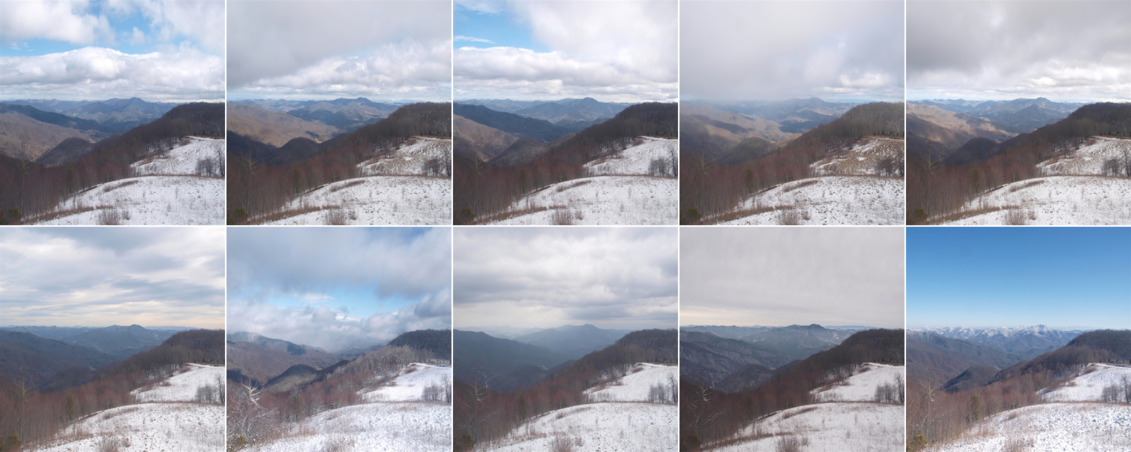}
    
  \end{tabular}
  
  \caption {Relating image appearance to the image context
    representation by visualizing the images that yield the highest
    activation at three different neurons.}

  \label{fig:activation}
\end{figure*}

\begin{figure}
  \centering
  
  \includegraphics[width=.98\textwidth]{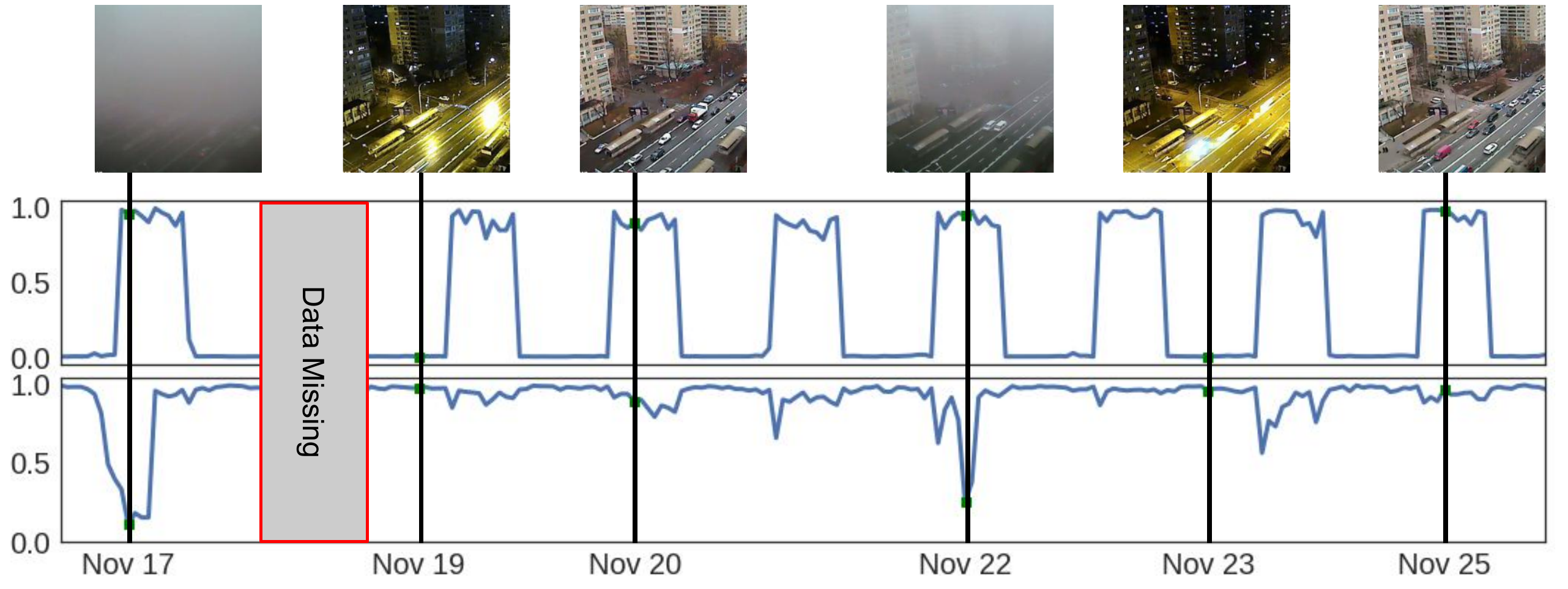}
  
  \caption{The time series of two neurons for a week of webcam
  imagery, with images showing the scene at various points. It appears
  that the top neuron is related to the diurnal cycle and the bottom
  is related to fogginess.}
  
  \label{fig:signal}
\end{figure}

We conducted several experiments to relate image appearance to the
representation learned by the image context network.
To begin, we examined images that correspond with extremal
activations.  For this experiment, we used 10,000 images randomly
sampled from the {\em YFCC} dataset and 7,732 images covering the year of 2015 from 
one webcam (ID: 4308) in the {\em AMOS} dataset. For each neuron of 
the image context representation, we selected the $10$ images that result in 
the highest activation from the two different sets of images.  \figref{activation} 
shows a montage of images for three neurons.  The neurons appear to capture
semantically meaningful attributes, such as daylight, rainy, and winter. 
Similarly, we selected two neurons and visualized their signal over
time for images from the webcam. \figref{signal} shows how
scene appearance changes are related to the image context features.
For the example shown, it appears that these neurons are related to
daylight and fogginess. 
These experiments provide evidence that the mid-level representation
captured by the image context network are related to static and
transient scene attributes.

\subsection{Analyzing Feature Correlation with Transient Attributes}

To analyze quantitatively how much our model learns about transient
attributes, we compute the cross correlations between a mid-level
representation of the image context network and the corresponding
transient attribute labels of all test images in the {\em TA} dataset.
As a baseline, we compare to features of the same architecture trained
for ImageNet~\cite{inception15} classification and features sampled
uniformly at random.  We select the feature
from the last pooling layer ({\em AvgPool\_1a\_7x7}), which is the
deepest layer that this model and ours share in common.
We compute the cross correlation scores between the feature and the
transient attribute scores of each image, resulting in a $1024 \times
40$ cross correlation matrix, $M$, where the element $m_{ij}$ is the
cross correlation score between the $i$-th feature channel and the
$j$-th transient attribute.
\figref{correlation} shows, for each
transient attribute, the maximum absolute correlation score
over all feature channels. We observe that our proposed method learns
features that are more correlated to the transient attributes
($\bar{\rho} = 0.414$) than the ImageNet network ($\bar{\rho} =
0.281$) or the random features ($\bar{\rho}=0.038$).

\begin{figure}
  \centering
  
  \includegraphics[width=\columnwidth]{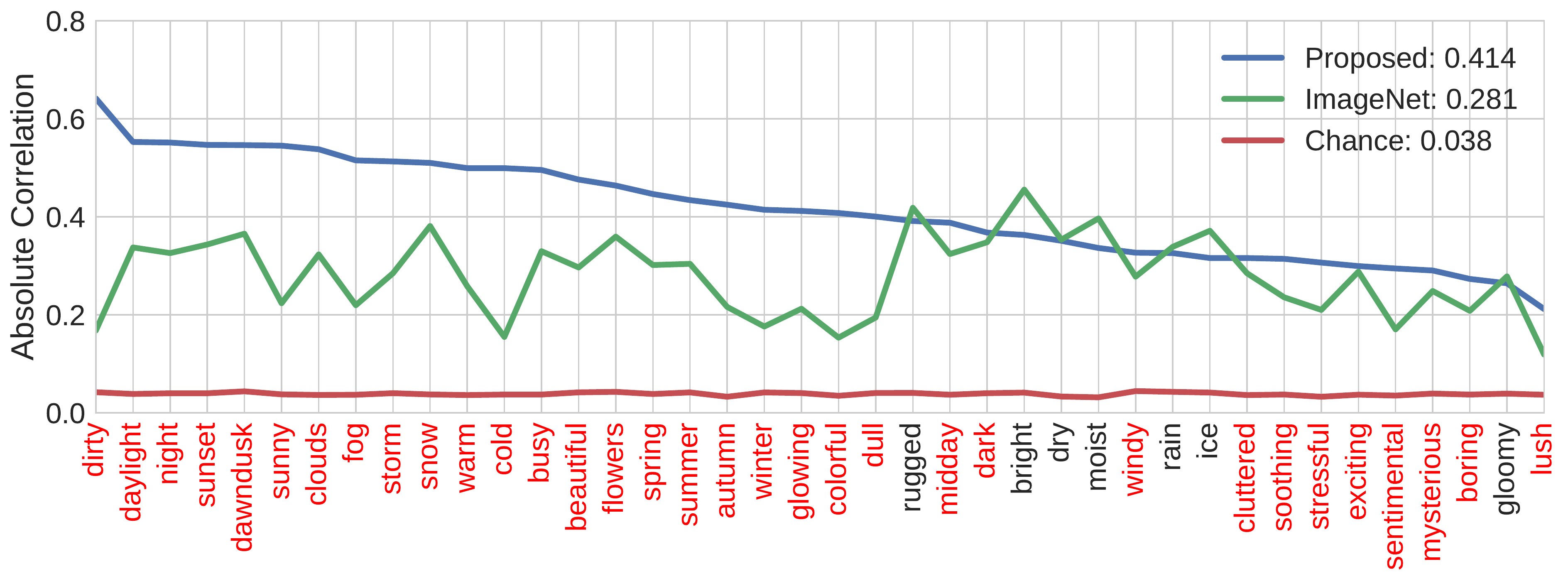} 
  
  \caption{Maximum absolute cross-correlation scores between transient attributes and
  three image feature representations. For a majority of attributes our proposed
  representation has at least one feature that is more highly correlated than any in both 
  baseline representations.}
  
  \figlab{correlation}
\end{figure}

\subsection{Comparing Mid-Level Features for Transient Attribute Estimation}

\begin{figure}
  \centering
  
  \includegraphics[width=.47\columnwidth]{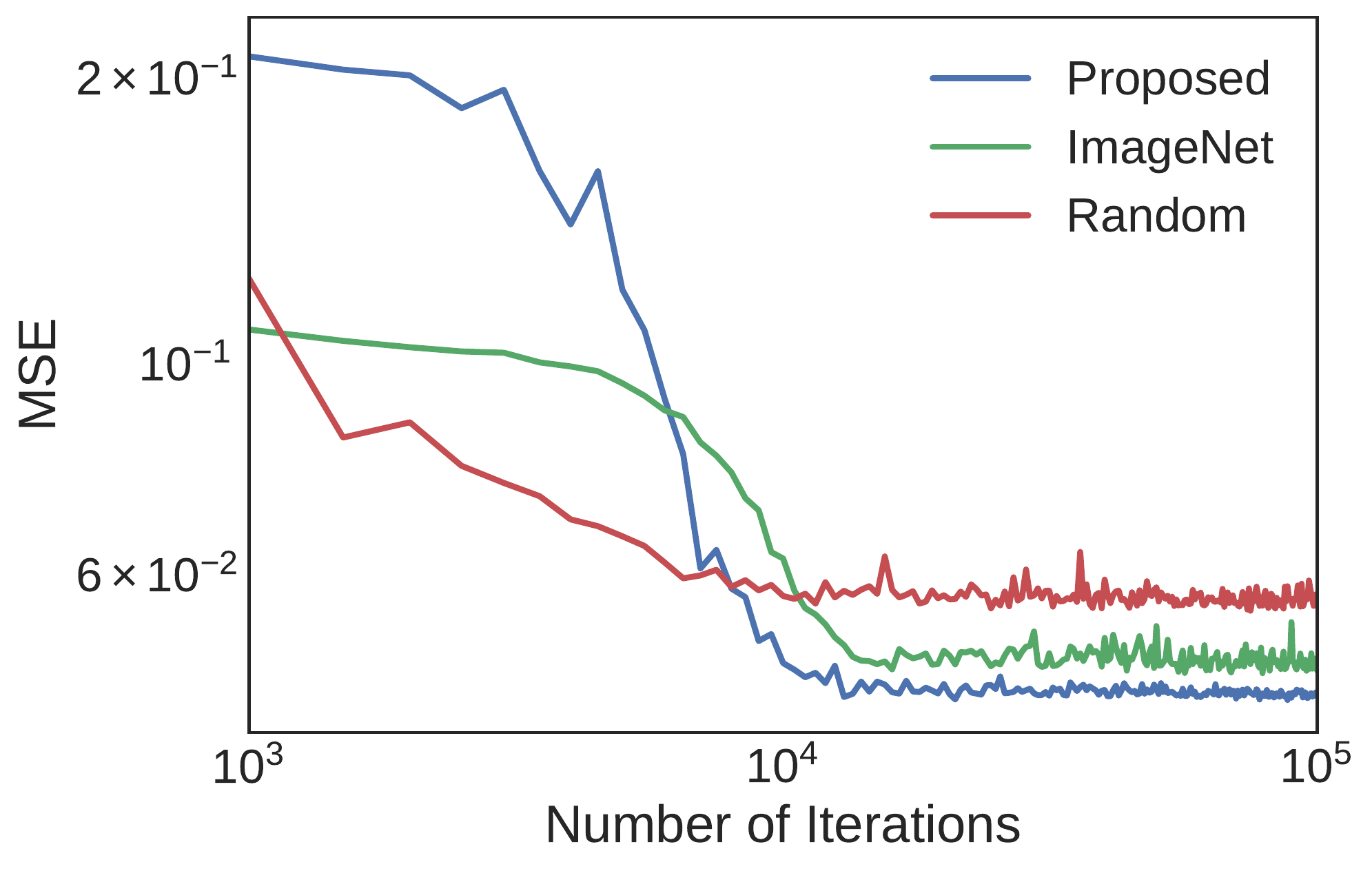}
  \includegraphics[width=.47\columnwidth]{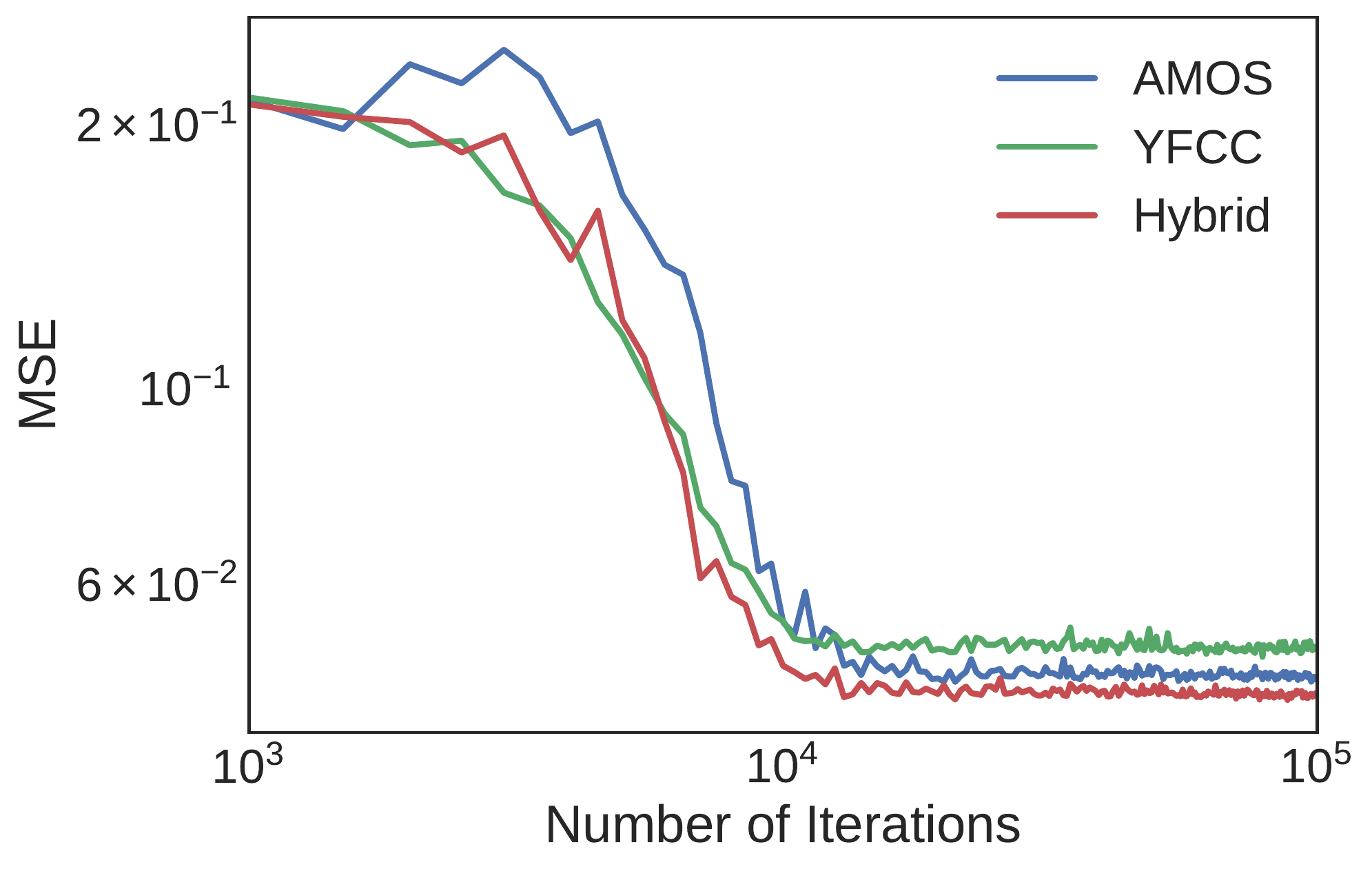}
  
  \caption{Comparing mid-level features for transient attribute
  estimation. (left) Features extracted from the weights of our proposed
  approach versus a network trained for image classification and a
  randomly initialized baseline. (right) Features extracted from our method,
  trained on different datasets.}
  
  \label{fig:pretrain}
\end{figure}

The previous experiment showed that the image context network is
capturing mid-level features correlated with transient attributes. In
this section, we explore the ability of this representation for
directly estimating transient attributes. Similar to the previous
experiment, we truncate our model at the last pooling layer (in order
to compare versus alternative initialization strategies), and add a
final two-layer MLP with 40 outputs corresponding to the 40 transient
attributes in the {\em TA} dataset. We train this
network, initializing from the weights of models trained for different
tasks, including variants of our method trained on the {\em AMOS}, {\em YFCC}, and
{\em Hybrid} datasets. During training, the MLP portions are randomly
initialized while the earlier layers are frozen. We evaluate the
average mean squared error (MSE) for the test set every 500 iterations
(batch size 32). \figref{pretrain} shows the performance comparison
among different mid-level features, including ImageNet and randomly
initialized {\em InceptionV2}. Our features are superior to
all baselines and perform best when learned using the {\em Hybrid} dataset.

\subsection{Application: Image Localization}

There are two image localization formulations that our network
architecture enables. The straightforward approach is to use the
location estimator (or the time-conditioned variant) to generate a
probability distribution over a discrete set of location bins. An
alternative approach is to optimize for a continuous location estimate
by minimizing the loss of the location-conditioned time estimator.

\begin{figure}
  \centering
  
  \begin{subfigure}[b]{0.47\linewidth}
    \includegraphics[width=\linewidth]{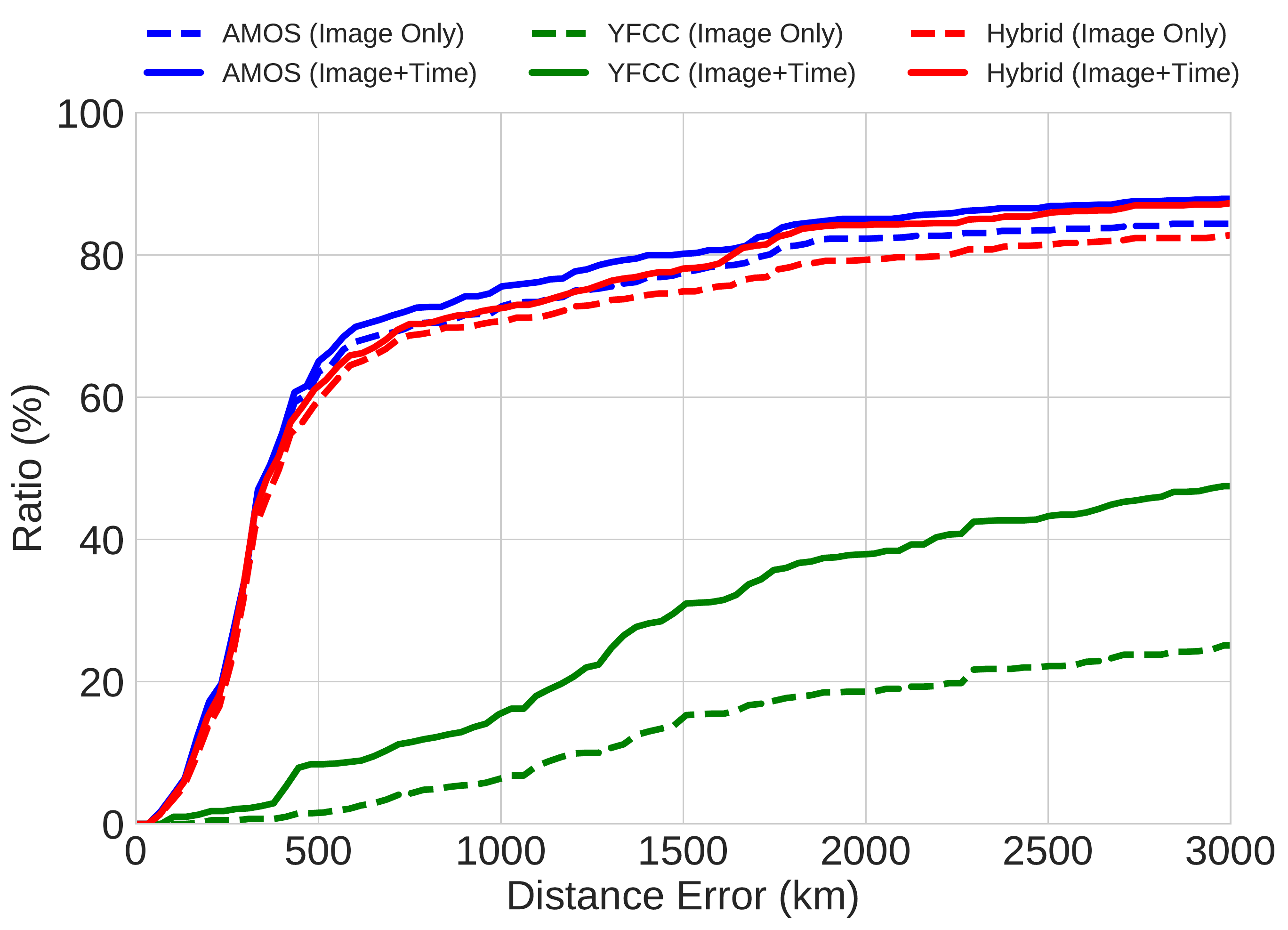}
    \caption{Test set: {\em AMOS}}
  \end{subfigure}
  \begin{subfigure}[b]{0.47\linewidth}
    \includegraphics[width=\linewidth]{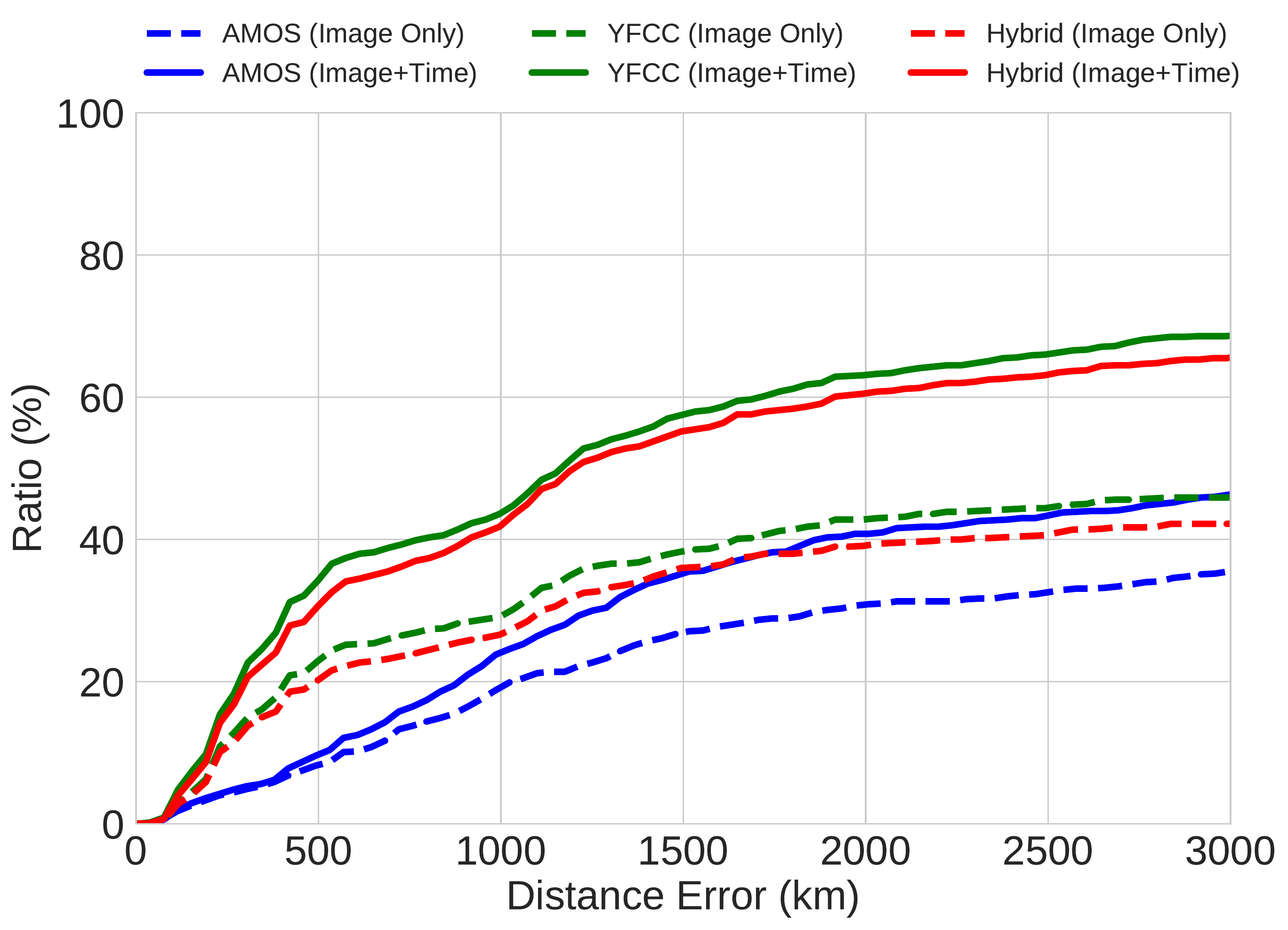}
    \caption{Test set: {\em YFCC}}
  \end{subfigure}
  
  \caption{Quantitative evaluation of localization performance shown
  as a cumulative distance error plot for the {\em YFCC} dataset.}
  
  \label{fig:image_localization}
\end{figure}

\paragraph{Discrete Localization} Given an input image, $I$, we
evaluate the location estimator $P(l|C_I(I))$ and the time-conditioned
location estimator $P(l|C_I(I),C_T(t))$, which requires a timestamp,
$t$. We trained our model on each dataset and perform quantitative
evaluation using the test images from {\em AMOS} and {\em YFCC},
separately. We use the latitude/longitude center of the highest probability bin as
our location estimate.  The results of this experiment are presented
in \figref{image_localization}.  We observe that the time-conditioned
location estimator is superior in both cases. We also conclude that
our model performs better if trained on the same imagery source with
the test set, and training the network with the {\em Hybrid} dataset
is competitive on both test sets. 

\begin{figure}
  \centering
  
  \setlength\tabcolsep{1pt}
  \renewcommand{\arraystretch}{0}
  \begin{tabular}{lcc}
    \raisebox{.5\height}{\rotatebox{90}{\em 1 images}} &
    \includegraphics[width=.45\linewidth]{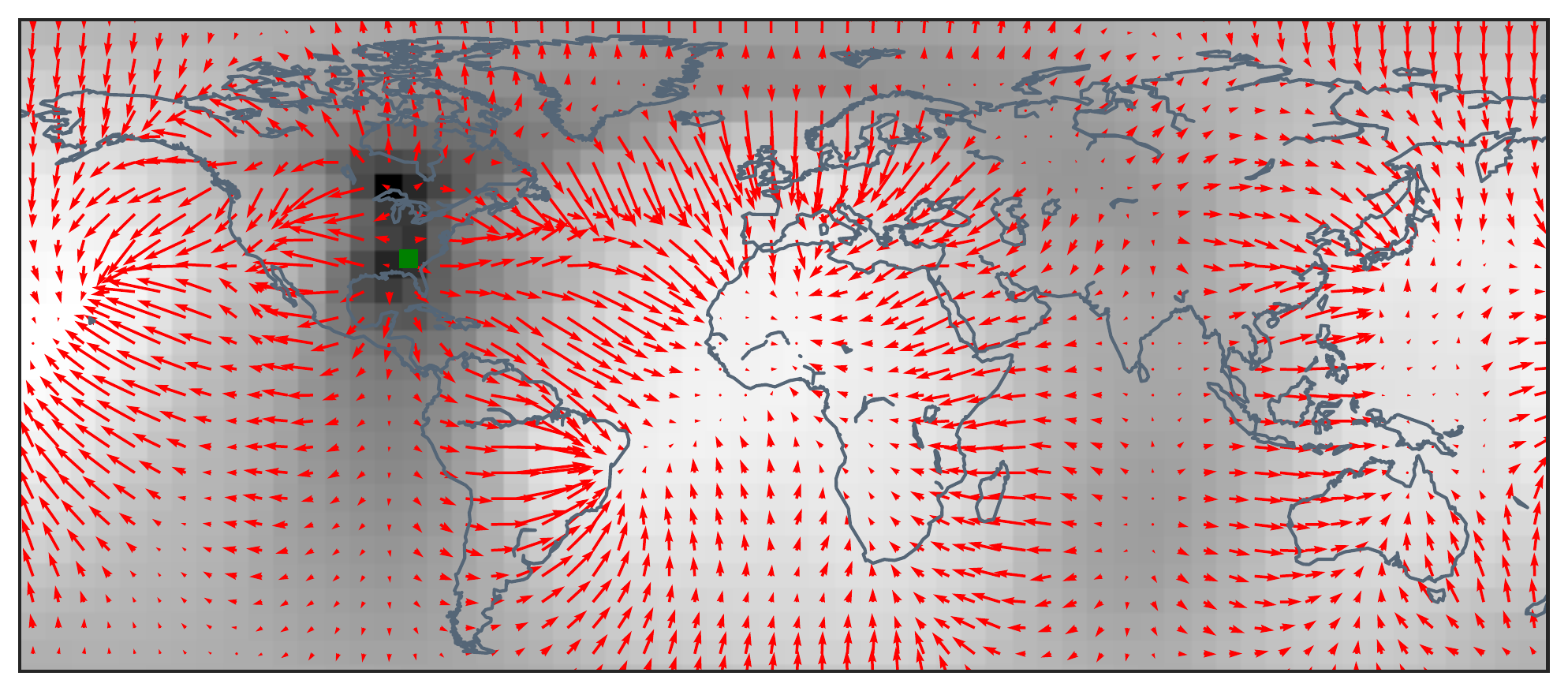} &
    \includegraphics[width=.45\linewidth]{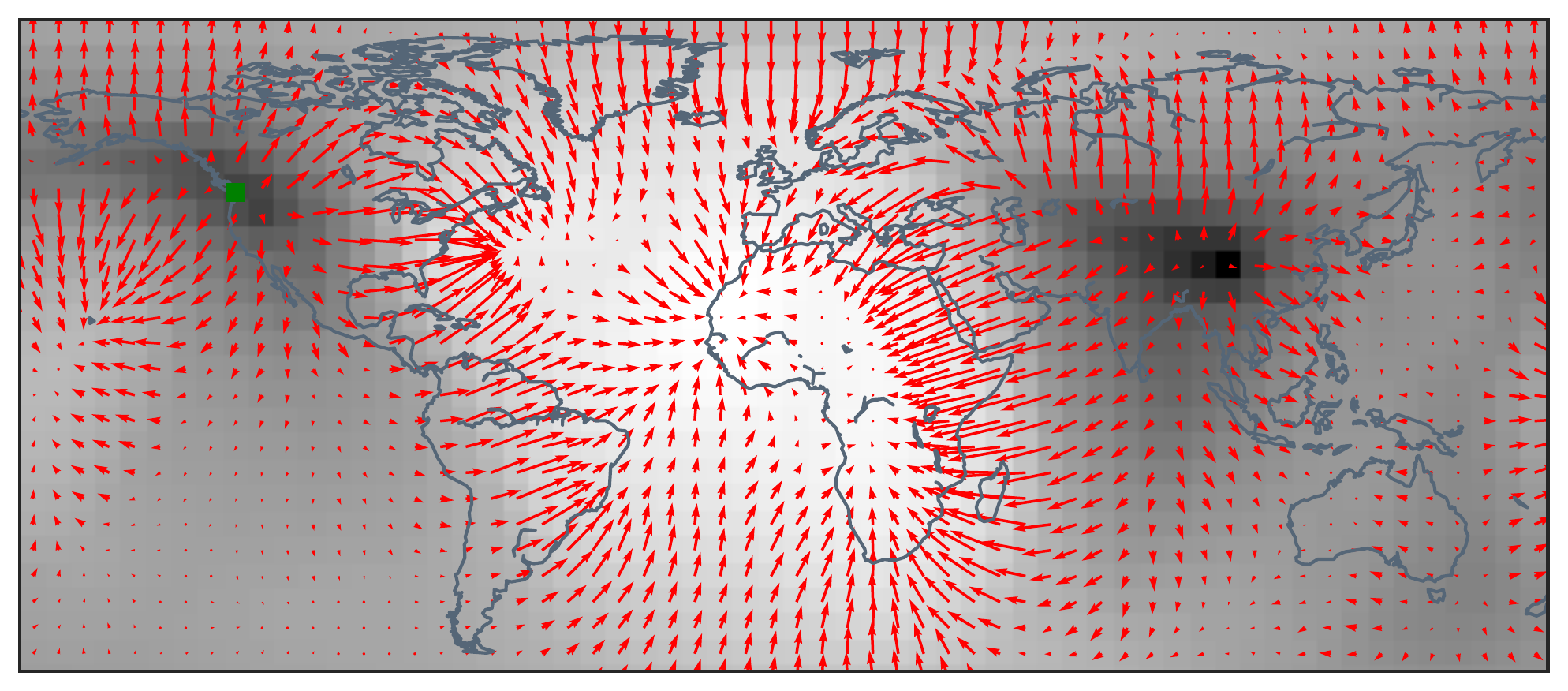} \\
    \vspace{.15cm}
    \raisebox{.5\height}{\rotatebox{90}{\em 5 images}} &
    \includegraphics[width=.45\linewidth]{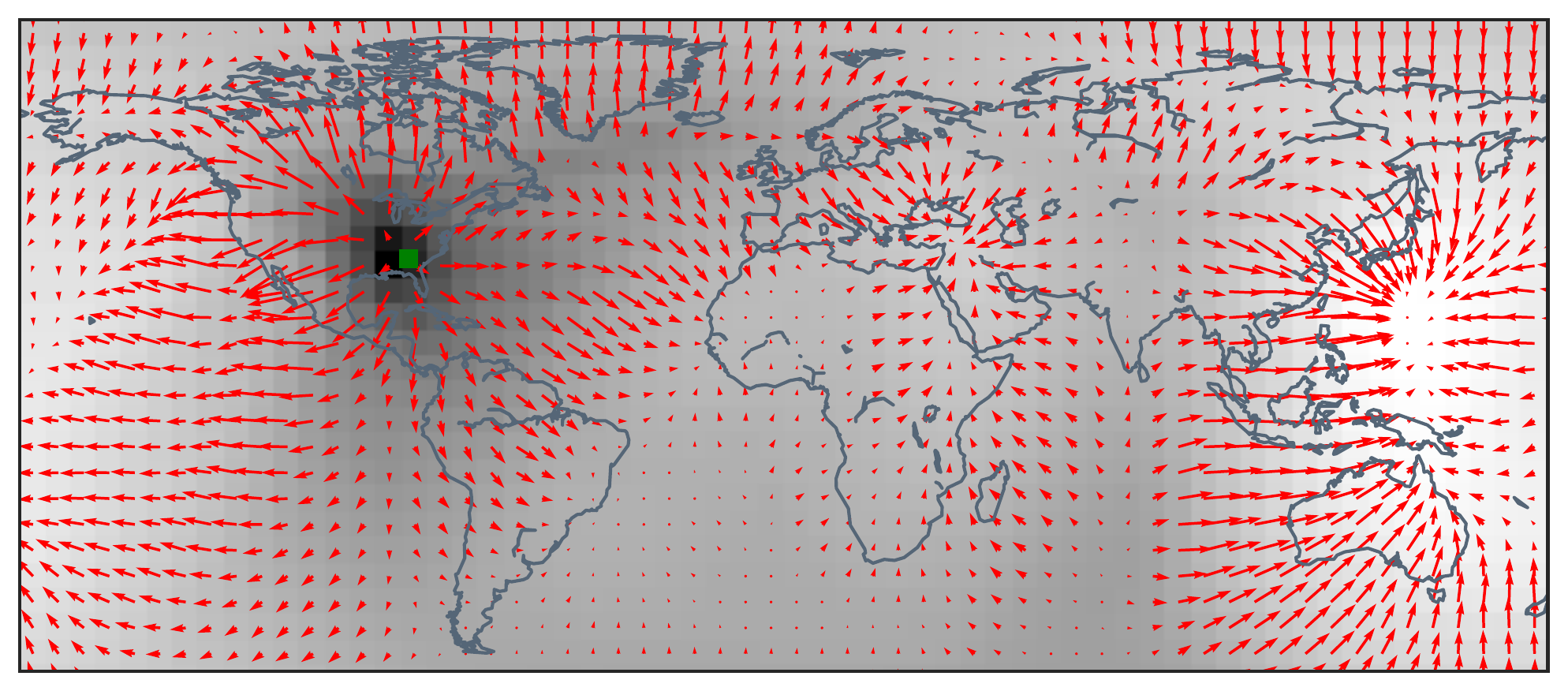} &
    \includegraphics[width=.45\linewidth]{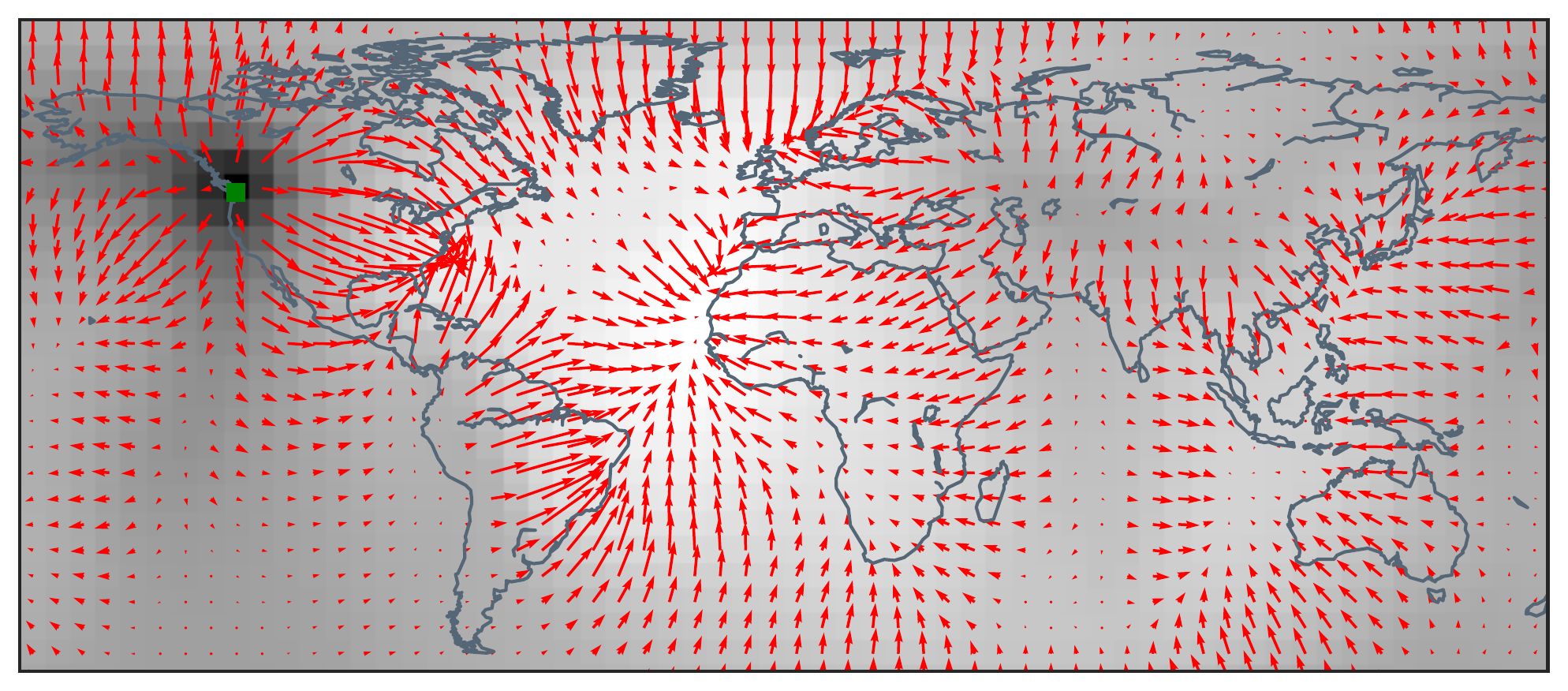} \\
    \vspace{.15cm}
    \raisebox{.5\height}{\rotatebox{90}{\em 20 images}} &
    \includegraphics[width=.45\linewidth]{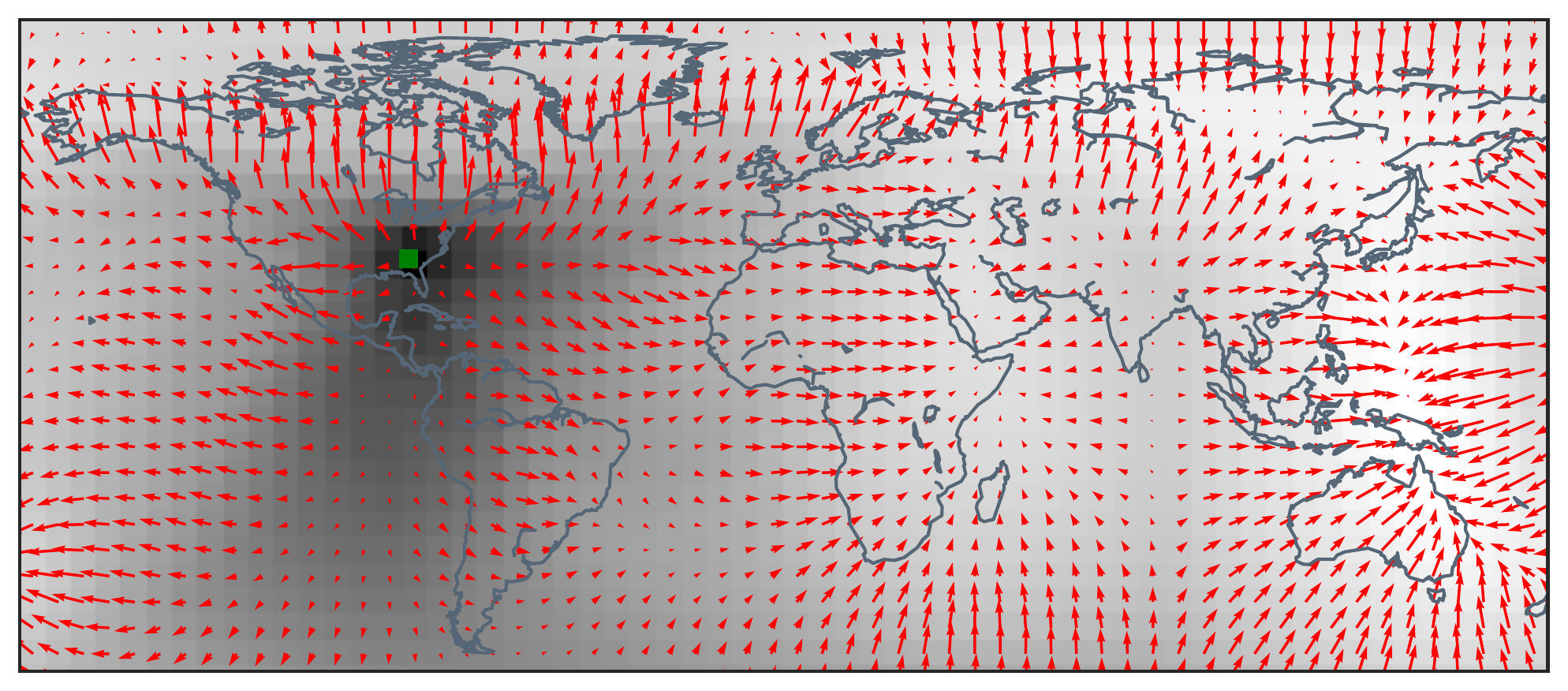} &
    \includegraphics[width=.45\linewidth]{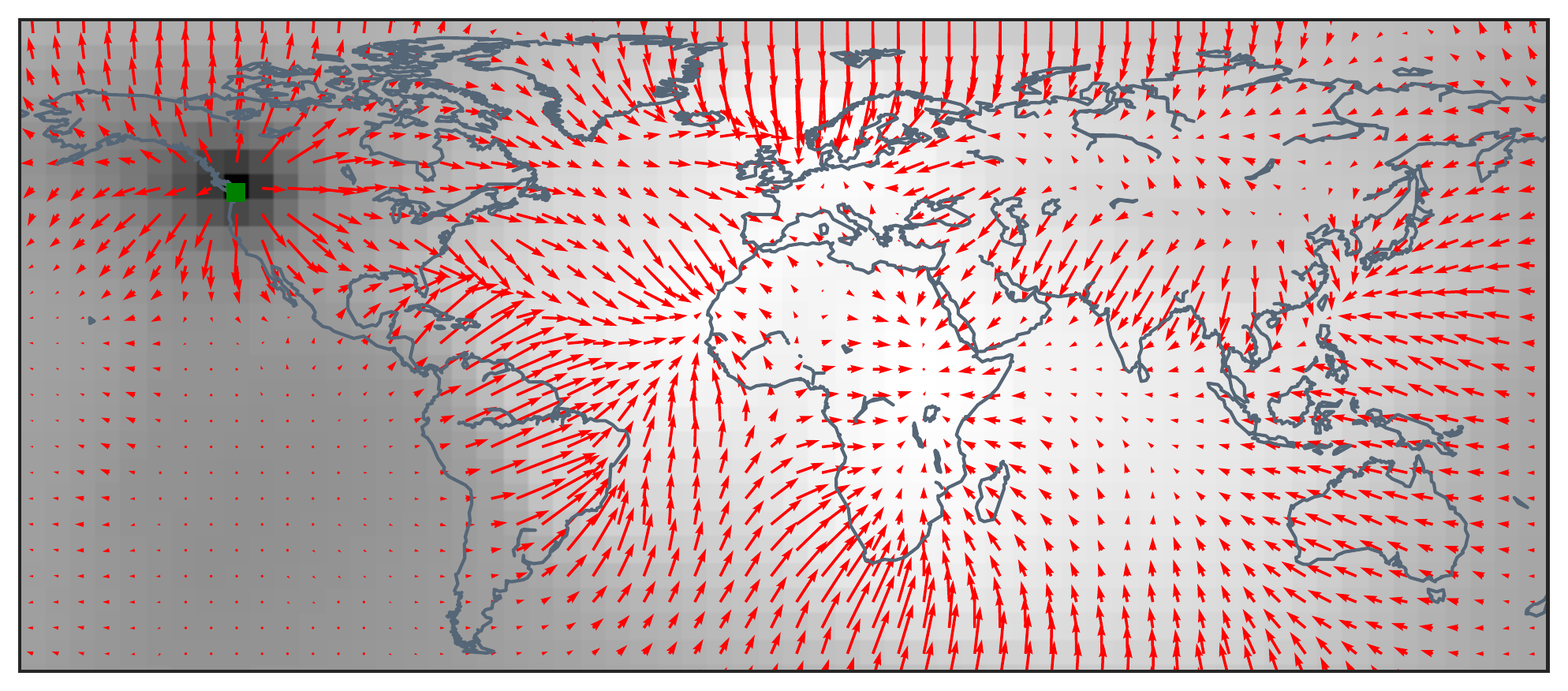} \\
    & {\em AMOS}: 5992 & {\em AMOS}: 8260 \\
  \end{tabular}
  
  \caption{Visualizing the time estimation loss for two webcams and
  varying number of images (darker is lower). The red arrows show the
  gradient and the green dot is the true location.}
  
  \label{fig:webcam_localization}
\end{figure}

\paragraph{Continuous Localization} In this formulation, we use
the location-conditioned time estimator, $P(t|C_I(I),C_L(l))$, to
optimize for a continuous location estimate.  Given the known image
capture time $t^*$, the idea is that the true location should result
in a low value for the loss associated with the estimator, $\ell_t =
\phi(P(t|C_I(I),C_L(l)), t^*)$, where $\phi(\circ,\circ)$ is the
cross-entropy loss. Therefore, we can produce a location estimate by
optimizing the location, $l$, with respect to $\ell_t$. Unfortunately,
an individual image does not typically yield a unique, or accurate,
location estimate using this method. However, if we sum the loss
across images captured at different times, we find that the minima of
the function becomes more distinct. \figref{webcam_localization} shows
several qualitative examples of this localization strategy on static
webcams, where darker colors correspond to more likely locations. We
can see that as additional images are included in the loss, the
uncertainty of the location prediction diminishes.

\subsection{Application: Time Estimation}

\begin{figure}
  \centering
  
  \begin{subfigure}[b]{0.45\linewidth}
    \includegraphics[width=\linewidth]{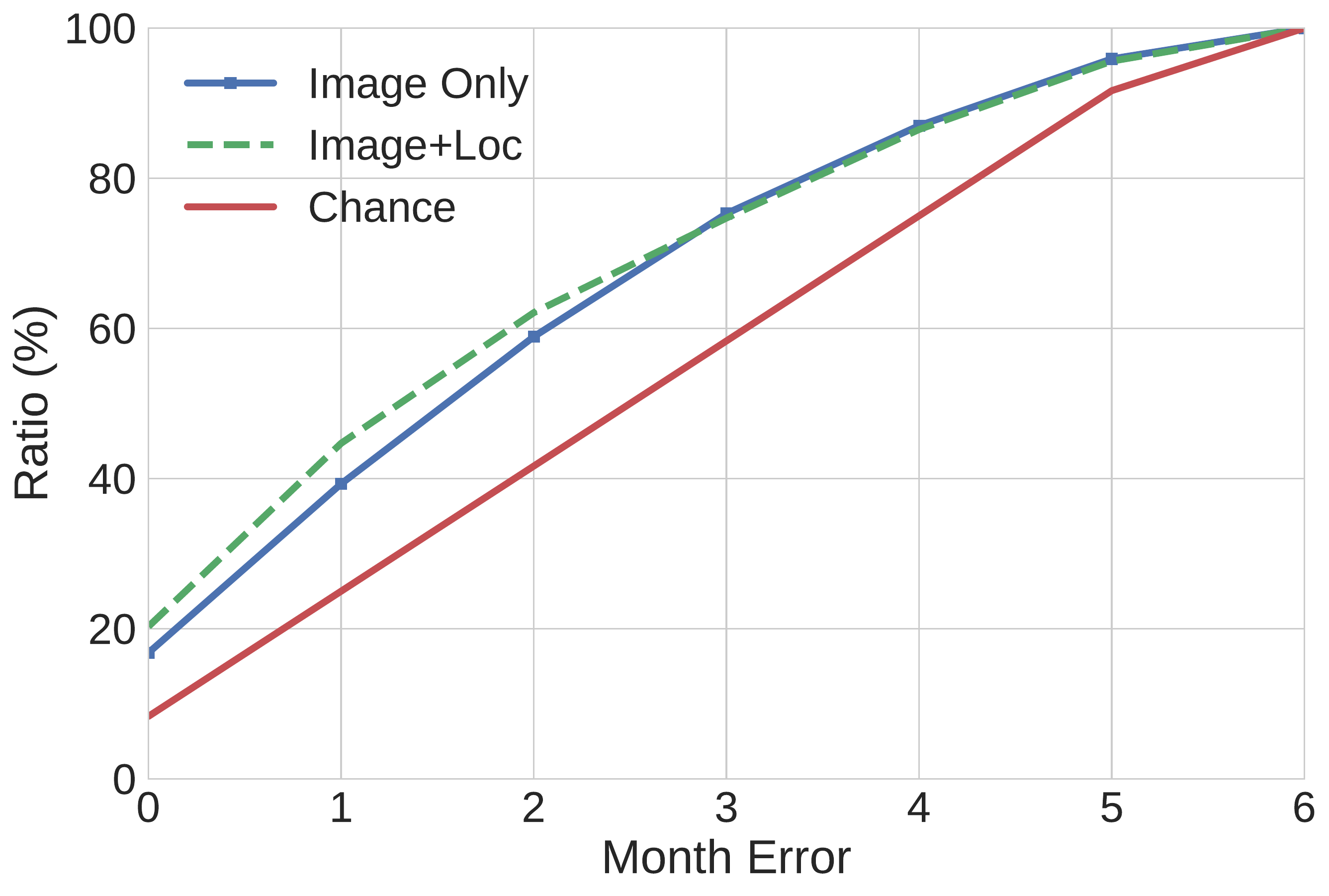}
    \caption{Month Differences}
  \end{subfigure}
  \begin{subfigure}[b]{0.45\linewidth}
    \includegraphics[width=\linewidth]{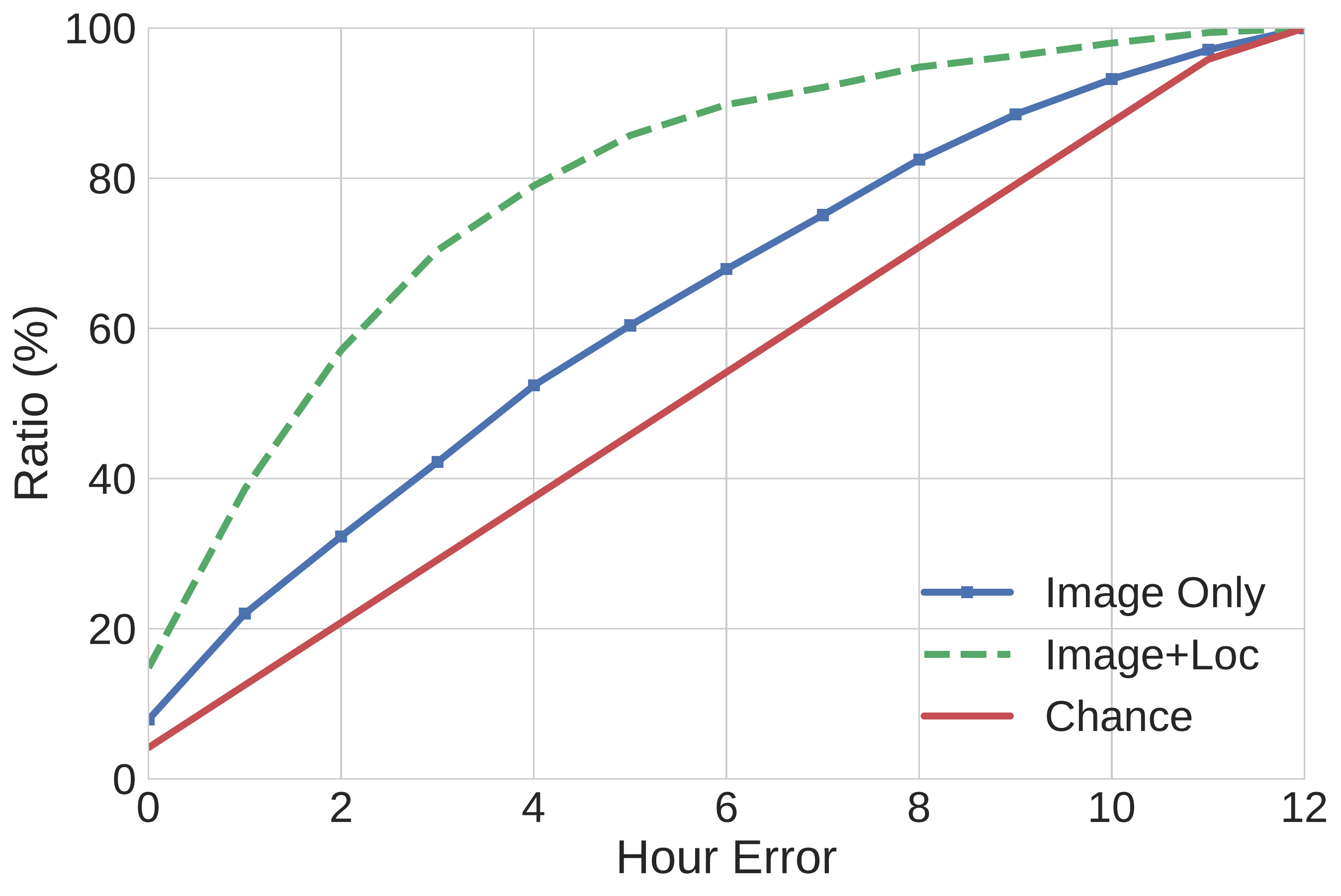}
    \caption{Hour Differences}
  \end{subfigure}
  
  \caption{Quantitative evaluation of marginal time estimation
  performance, shown as cumulative error plots for the {\em YFCC}
  dataset. Both methods perform better than the random chance and
  including the known location results in a significant reduction in
  error for the hour-estimation task.}
  
  \label{fig:time_diff_err}
\end{figure}

\begin{figure}
  \centering
  
  \begin{subfigure}[b]{.215\linewidth}
    \includegraphics[width=\linewidth]{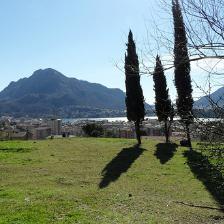}
    \caption{Input Image}
  \end{subfigure}
  \hspace{.25cm}
  \begin{subfigure}[b]{.33\linewidth}
    \includegraphics[width=\linewidth]{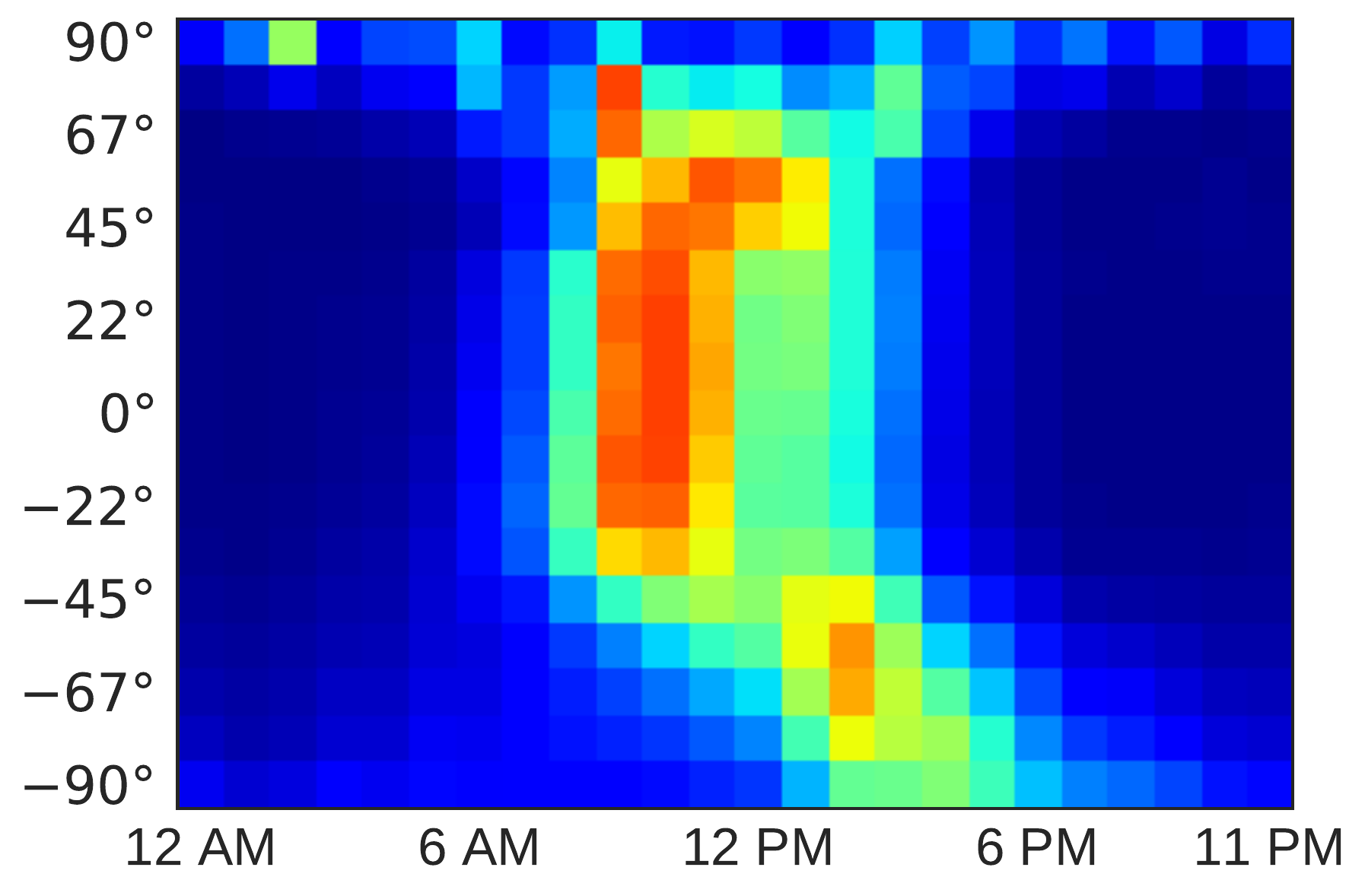}
    \caption{Latitude Changes}
  \end{subfigure}
  \begin{subfigure}[b]{.338\linewidth}
    \includegraphics[width=\linewidth]{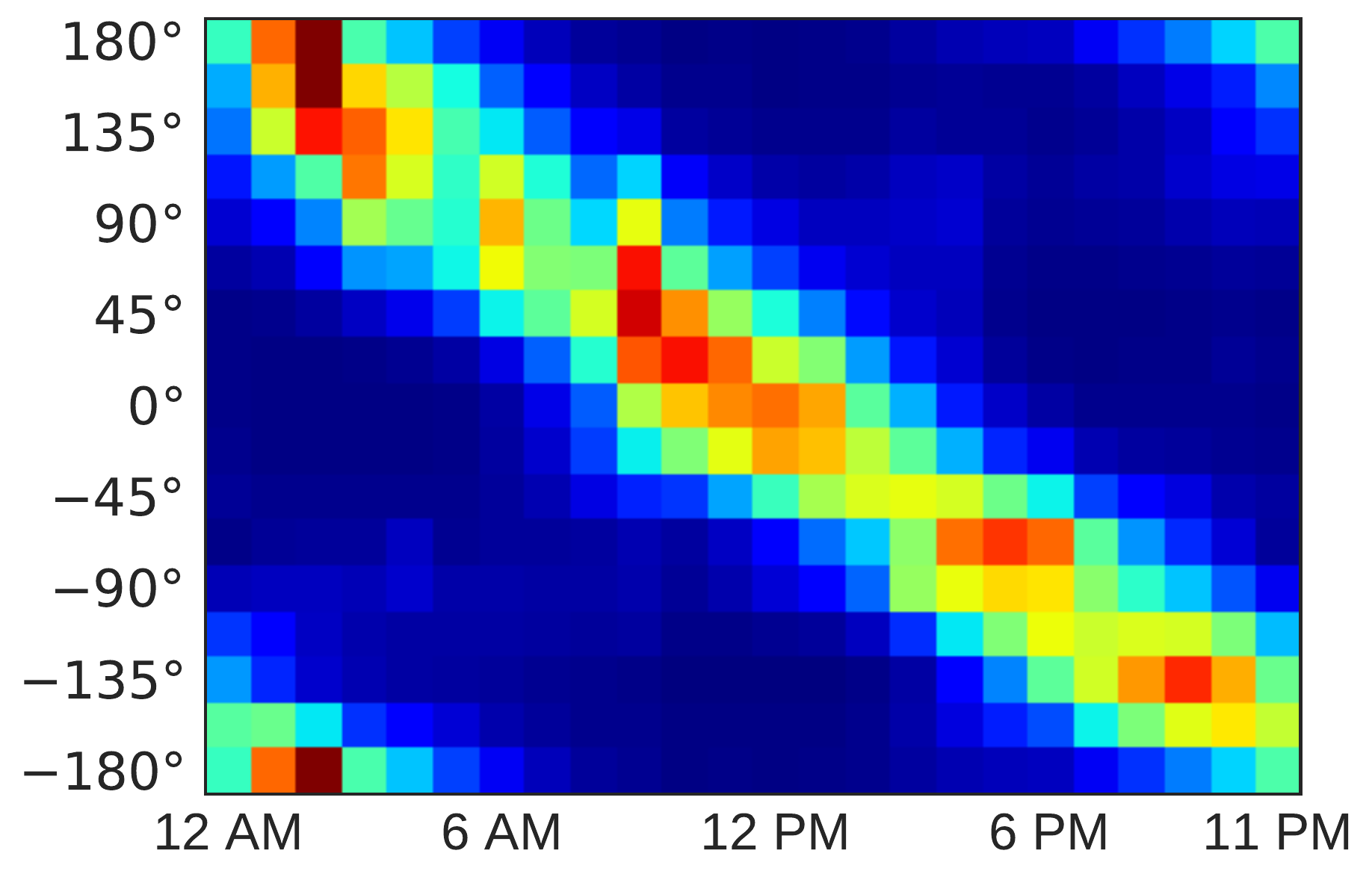}
    \caption{Longitude Changes}
  \end{subfigure}

  \caption{When using our location-conditioned time estimator, the
  marginal probability over hours changes significantly as we vary the
  latitude and longitude provided to the location context network.}
  
  \label{fig:hour-change}
\end{figure}

Using the time estimator and location-conditioned time estimator, our
network is able to estimate the capture time of a query image. These
estimators output a distribution in discrete 2D time space. To
evaluate our estimates, we compare the ground-truth capture time and
the marginal probabilities of our predictions on the {\em YFCC}
test set, and present the cumulative error plots in
\figref{time_diff_err}. We observe that including location is not
useful for  pinpointing the month. We suspect this is because most of
our imagery is in the northern hemisphere, and changing the location
within a hemisphere doesn't change the season. However, this is not
the case when estimating the hour.  To visualize this, we show in
\figref{hour-change} the impact of changing the location on the hour
estimate.  We compute the marginal hour distribution at different
latitudes and longitudes. When performing a sweep over latitude, we
fix the longitude value to be the ground truth (and vice versa). We
found that the longitude of the image corresponds more with the hour
prediction than the latitude, which matches expectations.

\section{Conclusion}

When learning about the world using images, the location and time an
image was captured are useful pieces of metadata that are often
available, but commonly overlooked. We presented a novel architecture
for learning useful representations from images that takes advantage
of this metadata. We found that for the task of transient attribute
estimation, our method, despite being trained without manually
obtained image-level annotations, learned image representations that
outperform the representations learned using ImageNet. This is a rarely
achieved feat in self-supervised representation learning against a
frequently used baseline.  One important area for future work is in
investigating alternative architectures for the context networks.  We
did not conduct a thorough study in this regard and expect to see
improvements in using newer image CNNs and higher capacity time and
location networks.  In addition, we expect that richer time and
location input representations will result in improved geo-temporal
image features.

\section*{Acknowledgement}
We gratefully acknowledge the support of NSF CAREER award IIS-1553116
and ARPA-E Award DE-AR0000594. The views and opinions of authors
expressed herein do not necessarily state or reflect those of the
United States Government or any agency thereof.

{
\small
\bibliography{biblio}

\begin{thebibliography}{32}
\providecommand{\natexlab}[1]{#1}
\providecommand{\url}[1]{\texttt{#1}}
\expandafter\ifx\csname urlstyle\endcsname\relax
  \providecommand{\doi}[1]{doi: #1}\else
  \providecommand{\doi}{doi: \begingroup \urlstyle{rm}\Url}\fi

\bibitem[Baltenberger et~al.(2016)Baltenberger, Zhai, Greenwell, Workman, and
  Jacobs]{baltenberger16transient}
Ryan Baltenberger, Menghua Zhai, Connor Greenwell, Scott Workman, and Nathan
  Jacobs.
\newblock {A Fast Method for Estimating Transient Scene Attributes}.
\newblock In \emph{IEEE Winter Conference on Applications of Computer Vision},
  2016.

\bibitem[Doersch(2016)]{doersch2016supervision}
Carl Doersch.
\newblock \emph{Supervision Beyond Manual Annotations for Learning Visual
  Representations}.
\newblock PhD thesis, Carnegie Mellon University, 2016.

\bibitem[Doersch et~al.(2015)Doersch, Gupta, and
  Efros]{doersch2015unsupervised}
Carl Doersch, Abhinav Gupta, and Alexei~A Efros.
\newblock Unsupervised visual representation learning by context prediction.
\newblock In \emph{IEEE International Conference on Computer Vision}, pages
  1422--1430, 2015.

\bibitem[Fedorov et~al.(2014)Fedorov, Fraternali, and
  Tagliasacchi]{fedorov2014snow}
Roman Fedorov, Piero Fraternali, and Marco Tagliasacchi.
\newblock Snow phenomena modeling through online public media.
\newblock In \emph{IEEE International Conference on Image Processing}, 2014.

\bibitem[Glorot and Bengio(2010)]{glorot2010understanding}
Xavier Glorot and Yoshua Bengio.
\newblock Understanding the difficulty of training deep feedforward neural
  networks.
\newblock In \emph{Proceedings of the thirteenth international conference on
  artificial intelligence and statistics}, pages 249--256, 2010.

\bibitem[Hays and Efros(2008)]{im2gps}
James Hays and Alexei~A. Efros.
\newblock im2gps: estimating geographic information from a single image.
\newblock In \emph{IEEE Conference on Computer Vision and Pattern Recognition},
  2008.

\bibitem[Ioffe and Szegedy(2015)]{inception15}
Sergey Ioffe and Christian Szegedy.
\newblock Batch normalization: Accelerating deep network training by reducing
  internal covariate shift.
\newblock In \emph{Proceedings of the 32Nd International Conference on
  International Conference on Machine Learning - Volume 37}, ICML'15, pages
  448--456. JMLR.org, 2015.

\bibitem[Islam et~al.(2013)Islam, Jacobs, Wu, and
  Souvenir]{islam13webcamweather}
Mohammad~T. Islam, Nathan Jacobs, Hui Wu, and Richard Souvenir.
\newblock Images+weather: Collection, validation, and refinement.
\newblock In \emph{IEEE CVPR Workshop on Ground Truth}, 2013.

\bibitem[Jacobs et~al.(2007{\natexlab{a}})Jacobs, Roman, and
  Pless]{jacobs07amos}
Nathan Jacobs, Nathaniel Roman, and Robert Pless.
\newblock Consistent temporal variations in many outdoor scenes.
\newblock In \emph{IEEE Conference on Computer Vision and Pattern Recognition},
  2007{\natexlab{a}}.

\bibitem[Jacobs et~al.(2007{\natexlab{b}})Jacobs, Satkin, Roman, Speyer, and
  Pless]{jacobs07geolocate}
Nathan Jacobs, Scott Satkin, Nathaniel Roman, Richard Speyer, and Robert Pless.
\newblock Geolocating static cameras.
\newblock In \emph{IEEE International Conference on Computer Vision},
  2007{\natexlab{b}}.

\bibitem[Jacobs et~al.(2013)Jacobs, Islam, and
  Workman]{jacobs13cloudcalibration}
Nathan Jacobs, Mohammad~T. Islam, and Scott Workman.
\newblock Cloud motion as a calibration cue.
\newblock In \emph{IEEE Conference on Computer Vision and Pattern Recognition},
  2013.

\bibitem[Kingma and Ba(2014)]{kingma2014adam}
Diederik~P Kingma and Jimmy Ba.
\newblock Adam: A method for stochastic optimization.
\newblock \emph{arXiv preprint arXiv:1412.6980}, 2014.

\bibitem[Laffont et~al.(2014)Laffont, Ren, Tao, Qian, and
  Hays]{laffont2014transient}
Pierre-Yves Laffont, Zhile Ren, Xiaofeng Tao, Chao Qian, and James Hays.
\newblock Transient attributes for high-level understanding and editing of
  outdoor scenes.
\newblock \emph{ACM Transactions on Graphics (SIGGRAPH)}, 33\penalty0
  (4):\penalty0 149, 2014.

\bibitem[Lalonde et~al.(2010)Lalonde, Narasimhan, and Efros]{lalonde2010sun}
Jean-Fran{\c{c}}ois Lalonde, Srinivasa~G Narasimhan, and Alexei~A Efros.
\newblock What do the sun and the sky tell us about the camera?
\newblock \emph{International Journal of Computer Vision}, 88\penalty0
  (1):\penalty0 24--51, 2010.

\bibitem[Lin et~al.(2015)Lin, Cui, Belongie, and Hays]{lin2015learning}
Tsung-Yi Lin, Yin Cui, Serge Belongie, and James Hays.
\newblock Learning deep representations for ground-to-aerial geolocalization.
\newblock In \emph{IEEE Conference on Computer Vision and Pattern Recognition},
  2015.

\bibitem[Lu et~al.(2014)Lu, Lin, Jia, and Tang]{lu2014two}
Cewu Lu, Di~Lin, Jiaya Jia, and Chi-Keung Tang.
\newblock Two-class weather classification.
\newblock In \emph{IEEE Conference on Computer Vision and Pattern Recognition},
  2014.

\bibitem[Matzen and Snavely(2014)]{matzen2014scene}
Kevin Matzen and Noah Snavely.
\newblock Scene chronology.
\newblock In \emph{European Conference on Computer Vision}, 2014.

\bibitem[Pathak et~al.(2016)Pathak, Krahenbuhl, Donahue, Darrell, and
  Efros]{pathak2016context}
Deepak Pathak, Philipp Krahenbuhl, Jeff Donahue, Trevor Darrell, and Alexei~A
  Efros.
\newblock Context encoders: Feature learning by inpainting.
\newblock In \emph{IEEE Conference on Computer Vision and Pattern Recognition},
  2016.

\bibitem[Pathak et~al.(2017)Pathak, Girshick, Doll{\'a}r, Darrell, and
  Hariharan]{pathak2017learning}
Deepak Pathak, Ross Girshick, Piotr Doll{\'a}r, Trevor Darrell, and Bharath
  Hariharan.
\newblock Learning features by watching objects move.
\newblock In \emph{IEEE Conference on Computer Vision and Pattern Recognition},
  2017.

\bibitem[Patterson and Hays(2012)]{patterson2012sun}
Genevieve Patterson and James Hays.
\newblock Sun attribute database: Discovering, annotating, and recognizing
  scene attributes.
\newblock In \emph{IEEE Conference on Computer Vision and Pattern Recognition},
  2012.

\bibitem[Russakovsky et~al.(2015)Russakovsky, Deng, Su, Krause, Satheesh, Ma,
  Huang, Karpathy, Khosla, Bernstein, Berg, and Fei-Fei]{ILSVRC15}
Olga Russakovsky, Jia Deng, Hao Su, Jonathan Krause, Sanjeev Satheesh, Sean Ma,
  Zhiheng Huang, Andrej Karpathy, Aditya Khosla, Michael Bernstein,
  Alexander~C. Berg, and Li~Fei-Fei.
\newblock {ImageNet Large Scale Visual Recognition Challenge}.
\newblock \emph{International Journal of Computer Vision}, 115\penalty0
  (3):\penalty0 211--252, 2015.
\newblock \doi{10.1007/s11263-015-0816-y}.

\bibitem[Salem et~al.(2016)Salem, Workman, Zhai, and Jacobs]{salem2016dating}
Tawfiq Salem, Scott Workman, Menghua Zhai, and Nathan Jacobs.
\newblock {Analyzing Human Appearance as a Cue for Dating Images}.
\newblock In \emph{IEEE Winter Conference on Applications of Computer Vision},
  2016.

\bibitem[Szegedy et~al.(2016)Szegedy, Vanhoucke, Ioffe, Shlens, and
  Wojna]{szegedy2016rethinking}
Christian Szegedy, Vincent Vanhoucke, Sergey Ioffe, Jon Shlens, and Zbigniew
  Wojna.
\newblock Rethinking the inception architecture for computer vision.
\newblock In \emph{IEEE Conference on Computer Vision and Pattern Recognition},
  2016.

\bibitem[Thomee et~al.(2015)Thomee, Shamma, Friedland, Elizalde, Ni, Poland,
  Borth, and Li]{yfcc100m}
Bart Thomee, David~A. Shamma, Gerald Friedland, Benjamin Elizalde, Karl Ni,
  Douglas Poland, Damian Borth, and Li{-}Jia Li.
\newblock Yfcc100m: The new data and new challenges in multimedia research.
\newblock \emph{CoRR}, abs/1503.01817, 2015.

\bibitem[Volokitin et~al.(2016)Volokitin, Timofte, and Gool]{eth_biwi_01292}
Anna Volokitin, Radu Timofte, and Luc~Van Gool.
\newblock Deep features or not: Temperature and time prediction in outdoor
  scenes.
\newblock In \emph{CVPR Workshop on Robust Features}, 2016.

\bibitem[Wang et~al.(2013)Wang, Korayem, and Crandall]{wang2013observing}
Jingya Wang, Mohammed Korayem, and David~J Crandall.
\newblock Observing the natural world with flickr.
\newblock In \emph{IEEE International Conference on Computer Vision Workshops},
  2013.

\bibitem[Weyand et~al.(2016)Weyand, Kostrikov, and Philbin]{weyand2016planet}
Tobias Weyand, Ilya Kostrikov, and James Philbin.
\newblock Planet-photo geolocation with convolutional neural networks.
\newblock In \emph{European Conference on Computer Vision}, pages 37--55.
  Springer, 2016.

\bibitem[Workman and Jacobs(2015)]{workman2015geocnn}
Scott Workman and Nathan Jacobs.
\newblock On the location dependence of convolutional neural network features.
\newblock In \emph{IEEE/ISPRS Workshop: Looking From Above: When Earth
  Observation Meets Vision}, 2015.

\bibitem[Workman et~al.(2014)Workman, Mihail, and Jacobs]{workman2014rainbow}
Scott Workman, R.~Paul Mihail, and Nathan Jacobs.
\newblock A pot of gold: Rainbows as a calibration cue.
\newblock In \emph{European Conference on Computer Vision}, 2014.

\bibitem[Workman et~al.(2015)Workman, Souvenir, and
  Jacobs]{workman2015localize}
Scott Workman, Richard Souvenir, and Nathan Jacobs.
\newblock Wide-area image geolocalization with aerial reference imagery.
\newblock In \emph{IEEE International Conference on Computer Vision}, pages
  1--9, 2015.

\bibitem[Zhang et~al.(2016)Zhang, Isola, and Efros]{zhang2016colorful}
Richard Zhang, Phillip Isola, and Alexei~A Efros.
\newblock Colorful image colorization.
\newblock In \emph{European Conference on Computer Vision}, 2016.

\bibitem[Zhou et~al.(2017)Zhou, Lapedriza, Khosla, Oliva, and
  Torralba]{zhou2017places}
Bolei Zhou, Agata Lapedriza, Aditya Khosla, Aude Oliva, and Antonio Torralba.
\newblock Places: A 10 million image database for scene recognition.
\newblock \emph{IEEE Transactions on Pattern Analysis and Machine
  Intelligence}, 2017.

\end{thebibliography}
}

\appendix

\end{document}